\documentclass[sigconf, screen=true]{acmart}

\usepackage{graphicx}
\usepackage{amsmath}
\usepackage{enumitem,kantlipsum}
\usepackage{multirow}
\usepackage{multicol}
\usepackage{xcolor}
\usepackage{subcaption}
\usepackage{mathtools}
\usepackage{csquotes}
\usepackage{color,soul}
\usepackage{arydshln}
\usepackage{bm}
\usepackage{hyperref}

\definecolor{rosegold}{rgb}{0.90, 0.70, 0.60}
\definecolor{mygreen}{rgb}{0.25, 0.67, 0.07}

\AtBeginDocument{%
  
  \hypersetup{linkcolor={red},citecolor={rosegold},urlcolor={black}}
  } 

\setcopyright{acmcopyright}
\copyrightyear{2023}
\acmYear{2023}
\acmDOI{XXXXXXX.XXXXXXX}

\acmConference[ICMR '23]{the 2023 International Conference on Multimedia Retrieval}{June 12--15,
  2023}{Thessaloniki, Greece}
\acmBooktitle{Proceedings of the 2023 International Conference on Multimedia Retrieval (ICMR '23), June 12--15, 2023, Thessaloniki, Greece.}
\acmPrice{15.00}
\acmISBN{978-1-4503-XXXX-X/18/06}

\begin{document}

\title{Exploration of Lightweight Single Image Denoising with Transformers and Truly Fair Training}

\author{Haram Choi}
\email{chlgkfka25@sogang.ac.kr}
\affiliation{%
  \institution{Machine Learning Lab., Sogang Univ.}
  \city{Seoul}
  \state{Mapo-gu}
  \country{Republic of Korea}
}

\author{Cheolwoong Na}
\email{ironyes@sogang.ac.kr}
\affiliation{%
  \institution{Machine Learning Lab., Sogang Univ.}
  \city{Seoul}
  \state{Mapo-gu}
  \country{Republic of Korea}
}

\author{Jinseop Kim}
\email{tjq2702@sogang.ac.kr}
\affiliation{%
  \institution{Machine Learning Lab., Sogang Univ.}
  \city{Seoul}
  \state{Mapo-gu}
  \country{Republic of Korea}
}

\author{Jihoon Yang}
\authornote{Corresponding author.}
\email{yangjh@sogang.ac.kr}
\affiliation{%
  \institution{Machine Learning Lab., Sogang Univ.}
  \city{Seoul}
  \state{Mapo-gu}
  \country{Republic of Korea}
}

\renewcommand{\shortauthors}{Choi et al.}

\begin{abstract}
As multimedia content often contains noise from intrinsic defects of digital devices, image denoising is an important step for high-level vision recognition tasks.
Although several studies have developed the denoising field employing advanced Transformers, these networks are too momory-intensive for real-world applications.
Additionally, there is a lack of research on lightweight denosing (LWDN) with Transformers.
To handle this, this work provides seven comparative baseline Transformers for LWDN, serving as a foundation for future research.
We also demonstrate the parts of randomly cropped patches significantly affect the denoising performances during training.
While previous studies have overlooked this aspect, we aim to train our baseline Transformers in a truly fair manner.
Furthermore, we conduct empirical analyses of various components to determine the key considerations for constructing LWDN Transformers.
Codes are available at \href{https://github.com/rami0205/LWDN}{https://github.com/rami0205/LWDN}.
\end{abstract}



\received{20 February 2007}
\received[revised]{12 March 2009}
\received[accepted]{5 June 2009}

\maketitle

\section{Introduction}

Since multimedia materials often contain noise generated by the intrinsic defect of sensor (\emph{e.g.}, in camera)~\cite{xu2018external,zha2019rank,tian2020deep}, image denoising (DN) is an important step before other downstream vision tasks.
Many convolutional neural networks (CNN) have improved this field~\cite{zhang2017beyond, zhang2018ffdnet, zhang2021plug, gou2022multi, renenhanced, shen2022adaptive}.
Meanwhile, after Vision Transformer (ViT)~\cite{dosovitskiy2020image} emerged, Transformers~\cite{vaswani2017attention} have substituted for CNNs in DN~\cite{chen2021pre,liang2021swinir, wang2022uformer, zamir2022restormer, zhang2022accurate, zheng2022cross, xiaostochastic}.
Nevertheless, it is infeasible to apply these models to real-world applications due to their intensive memory consumption (the number of parameters).
Unlike the lightweight super-resolution (SR) Transformers actively explored~\cite{choi2022n,fang2022hybrid,lu2021efficient,zhang2022efficient, du2022fast}, the most of lightweight DN (LWDN) studies~\cite{li2021lightweight, jia2021exploring, guo2021fast, zhou2022thunder, tian2023multi} have still adhered to conventional CNNs.
This unexplored field has few elaborate baselines to be compared with.
The well-designed baselines, however, are very important to provoke future works.
For instance, the lightweight SR studies began to be actively examined, only after several years (2016-2018) of monumental baselines proposed~\cite{kim2016accurate, kim2016deeply, lai2017deep, tai2017image, tai2017memnet, ahn2018fast, hui2018fast}.
Motivated by this, we carefully work on constructing and analyzing LWDN Transformer baselines\footnote{Compared with large DN models composed of 10M$\sim$50M parameters, we let LWDN models have around or below 1M parameters (4MB) following Choi et al.'s work~\cite{choi2022n}.}.

\begin{figure}[t]
    \centering
    \includegraphics[width=\linewidth]{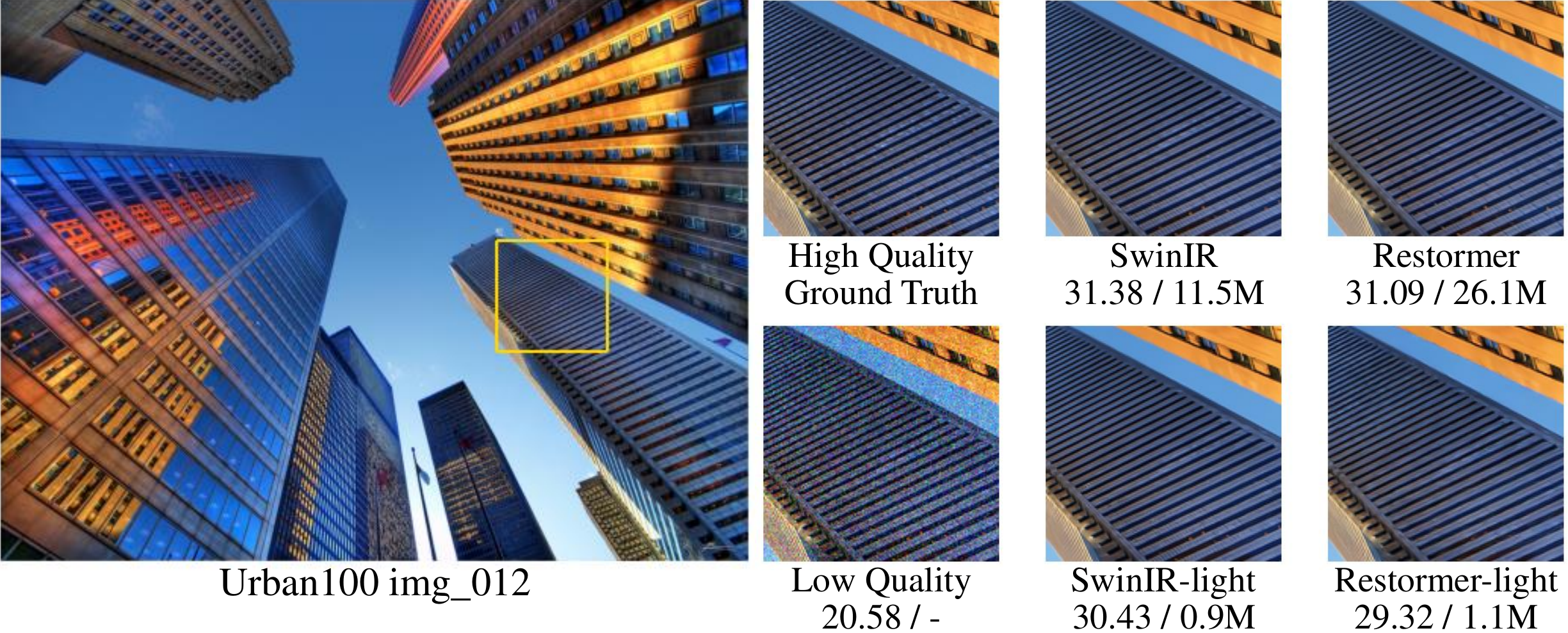}
    \caption{Visual examples of large and lightweight denosing results. PSNR (dB) / the number of parameters are compared. Although our lightweight baselines quantitatively fall behind the large counterparts due to much fewer weights, ours can recover similar texture to the large models for human-perception.}
    \label{fig_intro}
\end{figure}

\textbf{First}, we not only make the state-of-the-art (SOTA) large DN Transformers compact but also transplant SOTA lightweight SR Transformers into LWDN field for diverse baselines.
Specifically, the four best large DN methods are downsized, such as Uformer~\cite{wang2022uformer}, Restormer~\cite{zamir2022restormer}, ART~\cite{zhang2022accurate}, and CAT~\cite{zheng2022cross}.
We adopt three SOTA lightweight SR methods, such as SwinIR-light~\cite{liang2021swinir}, ELAN-light~\cite{zhang2022efficient}, and NGswin~\cite{choi2022n}.
These well-made Transformers, proposed during the last two years, represent our interesting field.
In terms of human-perception, they show comparable results to the large DN models with even fewer parameters, as illustrated in Figure~\ref{fig_intro}.

\textbf{Second}, we identify some unfairness in existing denoising studies.
We figure out an issue opposing to conventional wisdom: The numerous trials would almost remove the performance differences resulting from randomness.
Instead, since randomly selected patches for training process can obviously change the results (Section~\ref{randomness}), the direct comparisons in previous papers are inevitably unfair.
Consequently, we authentically control the randomness for training all models.
The same random patch from a training image is used by all networks at a certain iteration.
Additionally, while some studies trained their models with constant variance for deciding Gaussian noise level~\cite{zhang2019residual, zhang2020residual, chen2021pre, liang2021swinir}, others employed blind (unknown) one~\cite{zhang2017beyond, zhang2017learning, zhang2018ffdnet, zhang2021plug}.
Yet, the models learned with constant one is good at restoring a single level of noise but bad at recovering the other noise levels.
Thus, we standardize our work by using blind noise level for training all models.

\textbf{Third}, we empirically analyze the different components of our baselines.
Please note that we do not present new methods to enhance the performances.
However, our novelty is that we establish the baselines for an under-explored topic, and deliver interpretability and insight, thereby encouraging future research.
Starting with a (1)\underline{hierarchical network}, we characterize it by three aspects: the encoder connection, bottleneck input, and decoder structure.
We apply the robust and advanced elements proposed by~\cite{choi2022n} with respect to these aspects to another hierarchical network, and confirm the potential of hierarchical structures to be improved.
Next, we discover that the (2)\underline{channel self-attention} is worse at recovering the noisy images than the spatial self-attention methods, under the parameter constraint (\emph{i.e.}, lightweight condition).
After that, we show (3)\underline{excessive weight sharing} may lead to unstable learning due to limited flexibility and representation of the network.
At last, we illuminate that the careful (4)\underline{design of CNNs} is still relevant in the present where self-attention is widely adopted by varying the shared tail module composed of only CNNs.

The summarized main contributions are as follows:
\vspace{-2pt}

\begin{enumerate}[label=(\arabic*), left=0.05cm]
    \item We provide various comparison groups of lightweight Transformer architectures for color and grayscale Gaussian denoising, which have not been explored until recently. Three lightweight super-resolution and four state-of-the-art large denoising methods are used to establish LWDN Transformer baselines. They can serve as foundation of active future studies (Sections~\ref{lwdn},~\ref{common},~\ref{main_results}).
    \item Since many image restoration papers have overlooked the truly same training settings, we aim to implement the authentically fair experiments. All models used in this paper are trained on identically cropped random patches (Sections~\ref{fair_training},~\ref{randomness}).
    \item Some empirical studies on different components provide interpretability or insight for LWDN field. These practices are expected to facilitate and inspire future works (Section~\ref{analysis}).
\end{enumerate}

\section{Related Work}

\textbf{Importance of Baselines.}
The models with remarkable improvements take several years to be accumulated so that the research area evolves independently.
For example, lightweight super-resolution (SR) had been a separate area, only after several years of monumental baselines proposed~\cite{kim2016accurate, kim2016deeply, lai2017deep, tai2017image, tai2017memnet, ahn2018fast, hui2018fast} (2016-2018).
Afterwards, many researchers introduced lightweight SR networks~\cite{hui2019lightweight,luo2020latticenet,liu2020residual,choi2022n,fang2022hybrid,lu2021efficient,zhang2022efficient, du2022fast}.
This phenomenon was also observed in other unrelated fields, such as reinforcement learning (RL).
After DQN~\cite{mnih2013playing} introduced a deep learning method in RL, various innovative methods were proposed over a few years (2015-2018)~\cite{lillicrap2015continuous,mnih2016asynchronous,schulman2017proximal,haarnoja2018soft}.
Since then, other deep learning approaches have been developed in RL~\cite{badia2020never,badia2020agent57}.
Meanwhile, well-designed lightweight SR and large DN Transformers have been proposed over the past two years.
Our work takes advantages of these techniques to shorten the periods for future LWDN research with Transformers.

\noindent
\textbf{Image Restoration.}
Many Transformer-based approaches improved image restoration (IR) performances, such as image denoising (DN) and super-resolution (SR).
SwinIR~\cite{liang2021swinir} exploited local window self-attention (SA)~\cite{vaswani2017attention} of Swin Transformer~\cite{liu2021swin}.
Subsequent studies focused on expanding the receptive field while leveraging the long-range dependencies of SA.
Uformer~\cite{wang2022uformer} introduced locally enhanced feed-forward network while keeping a U-Net structure~\cite{ronneberger2015u}.
Restormer~\cite{zamir2022restormer} performed global SA in a channel space instead of spatial dimension.
ELAN~\cite{zhang2022efficient} employed shift-convolution~\cite{wu2018shift} and multi-scaled local window SA.
CAT~\cite{zheng2022cross} replaced a square window with a rectangular one.
ART~\cite{zhang2022accurate} introduced sparse attention by dilated window SA.
NGswin~\cite{choi2022n} proposed N-Gram embedding that considers neighboring regions of each window before SA.

\noindent
\textbf{Patch-Driven IR.}
Our attempt at fair training is related to interpretation studies.
They implied that the patches selected for training should be deemed important.
As prior work, the authors of~\cite{gu2021interpreting} proposed a local attribution map (LAM) to visualize the contribution of each pixel in image recovery.
They demonstrated that some areas in a local patch, like edges and textures, significantly affect the restoration performances.
Magid et al.~\cite{magid2022texture} evaluated the error based on semantic labels from a learned texture-classifier.
They distinguished between more complex and simpler textures of low-quality images to restore.
The researchers of RCAN-it~\cite{lin2022revisiting} hypothesized that if a network were trained more on the low-quality patches that have a lower PSNR over their high-quality counterparts, the performance could be improved.
Although the performances decreased, they found that there were attributes of the random patches that influence the low-level vision tasks.
In spite of those evidences, existing IR papers have overlooked the influences of randomly selected patches and compared their works in an unfair manner.

\begin{table*}[t]
    \centering
    \caption{Summary of the characteristics of our lightweight denoising baseline Transformers. \enquote{Hier.} indicates whether each network adopts a hierarchical U-Net~\cite{ronneberger2015u} based architecture or a non-hierarchical structure.} 
    \label{tab_summary}
    \resizebox{0.6\linewidth}{!}
    {\begin{tabular}{l|c|l|l|c}
    \hline
    Method & Hier. & Self-attention (SA) & Feed-forward network & Bottleneck \\
    \hline
    SwinIR-light & X & Plain window~\cite{liu2021swin} & Plain~\cite{dosovitskiy2020image} & - \\
    ELAN-light & X & Multi-scale window & Before SA, Shift-conv~\cite{wu2018shift} & - \\
    NGswin & O & N-Gram neighbor window & Post-layer-norm~\cite{liu2022swin} & SCDP \\
    Restormer-light & O & Channel space~\cite{zhang2018image} & Adding depthwise conv & Transformer \\
    Uformer-light & O & Plain window~\cite{liu2021swin} & Adding depthwise conv & Transformer \\
    CAT-light & O & Rectangle window & Plain~\cite{dosovitskiy2020image} & Transformer \\
    ART-light & X & Sparse and dense window & Plain~\cite{dosovitskiy2020image} & - \\
    \hline
    \end{tabular}}
\end{table*}

\section{Methodology}

\subsection{LWDN Transformer}
\label{lwdn}

Employing seven state-of-the-art Transformer methods, we establish baselines for lightweight denoising (LWDN).
Three models originate from lightweight super-resolution task, including SwinIR-light~\cite{liang2021swinir}, ELAN-light~\cite{zhang2022efficient}, and NGswin~\cite{choi2022n}.
Each architecture remains unchanged, with an exception of the final reconstruction module (See Section~\ref{common}).
The other four Transformers come from the large DN task, including Restormer~\cite{zamir2022restormer}, Uformer~\cite{wang2022uformer}, CAT~\cite{zheng2022cross}, and ART~\cite{zhang2022accurate}.
We reduce the number of Transformer blocks and channels, or change other hyper-parameters.
As a result, the total number of learnable parameters in each model is set to around 1M.
The details of reductions are in Table\textcolor{red}{~\ref{tab_reduction}}.
We also summarize the attributes of the network components in each model in Table~\ref{tab_summary}.

\begin{table}
    \centering
    \caption{Reduction of large to lightweight DN. \enquote{Depth} indicates the number of Transformer blocks in each layer. \enquote{Hidden (FFN)} means the hidden dimension in feed-forward network after self-attention. We keep the number of learnable parameters as around one million.}
    \label{tab_reduction}
    \resizebox{\linewidth}{!}
    {\begin{tabular}{|c|c|c|c|c|}
    \hline
    Model & Depth & Channels & Hidden (FFN) & \#Params \\
    \hline
    \multirow{2}{*}{Restormer~\cite{zamir2022restormer}} & [4, 6, 6, 8, 6, 6, 4, 4] $\rightarrow$ & \multirow{2}{*}{48 $\rightarrow$ 16} & \multirow{2}{*}{128 $\rightarrow$ 32} & \multirow{2}{*}{26,112K $\rightarrow$ 1,054K} \\
    & [2, 2, 2, 2, 2, 2, 2, 2] & & & \\
    \hline
    \multirow{2}{*}{Uformer~\cite{wang2022uformer}} & [1, 2, 8, 8, 2, 8, 8, 2, 1] $\rightarrow$ & \multirow{2}{*}{32 $\rightarrow$ 16} & \multirow{2}{*}{128 $\rightarrow$ 32} & \multirow{2}{*}{50,881K $\rightarrow$ 1,084K} \\
    & [2, 4, 2, 2, 2, 4, 2] & & & \\
    \hline
    \multirow{2}{*}{CAT~\cite{zheng2022cross}} & [4, 6, 6, 8, 6, 6, 4, 4] $\rightarrow$ & \multirow{2}{*}{48 $\rightarrow$ 16} & \multirow{2}{*}{128 $\rightarrow$ 32} & \multirow{2}{*}{25,770K $\rightarrow$ 1,042K} \\
    & [2, 2, 4, 2, 4, 2, 2, 2] & & & \\
    \hline
    \multirow{2}{*}{ART~\cite{zhang2022accurate}} & [6, 6, 6, 6, 6, 6] $\rightarrow$ & \multirow{2}{*}{180 $\rightarrow$ 60} & \multirow{2}{*}{720 $\rightarrow$ 120} & \multirow{2}{*}{16,150K $\rightarrow$ 1,084K} \\
    & [6, 6, 6, 6, 6] & & & \\
    \hline
    \end{tabular}}
\end{table}

\subsection{Shared Common Components}
\label{common}

\begin{figure}[t]
    \centering
    \includegraphics[width=\linewidth]{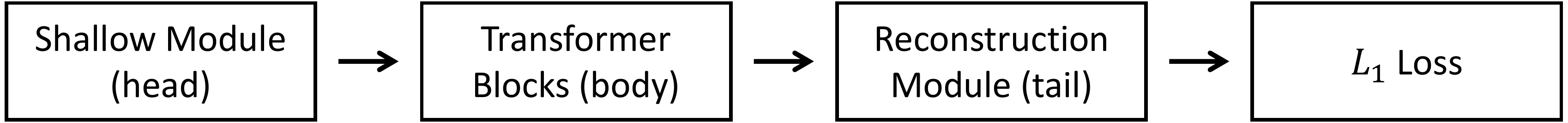}
    \caption{Brief pipeline of baselines. The only difference between each model is Transformer block (body).}
    \label{fig_brief}
    \vspace{-10pt}
\end{figure}

To maintain consistency across models, we apply identical shallow (or head) module, reconstruction (or tail) modules, and loss function to all models.
Figure~\ref{fig_brief} depicts the brief pipeline.
The only difference is the Transformer blocks (body).
This unity assures to identify the effectiveness of unique algorithms in self-attention and feed-forward networks, which are the key factors of Transformers.

\noindent
\textbf{Shallow Module.}
This module consists of a $3\times3$ convolution.
It takes a low-quality noisy image $I_{LQ} \in \mathbb{R}^{C_{in} \times H\times W}$, extracting the shallow feature $z_s \in \mathbb{R}^{C \times H \times W}$,
where $C_{in}$ is 1 or 3 according to whether grayscale or color input, and H and W indicate the resolution of the input.
$C$ is the embedding dimension (channels) of each network.

\noindent
\textbf{Reconstruction Module.}
The final reconstruction module $\mathcal{F}_{recon}$ is composed of two $3\times3$ convolutional layers.
The first adjusts the channels of feature maps to $C_{out}$, which is equal to $C_{in}$.
Then the second layer produces the residual output $I_{res}$, which is added to $I_{LQ}$.
Finally, we get the reconstructed clean image $I_{RC}$, as follows:
\begin{equation}
    \label{eq_recon}
    \begin{aligned}
        I_{res} = \mathcal{F}_{recon}(\mathcal{F}_{body}(z_s)),\; I_{RC} = I_{LQ} + I_{res},
    \end{aligned}
\end{equation}
where $\mathcal{F}_{body}$ represents the Transformer blocks.
The tail modules of SwinIR-light, ELAN-light, and NGswin differ from the original ones.
An upsampling pixel-shuffle~\cite{shi2016real} layer is removed.
In Section~\ref{cnnstill}, we examine the variants of this module.
This is because image restoration tasks still need convolution for aggregating local features despite the robustness of self-attention~\cite{zheng2022cross}.

\noindent
\textbf{Loss Function.}
We minimize $L_1$ pixel loss for training LWDN baseline networks: $\mathcal{L} = \lVert I_{HQ}-I_{RC} \rVert_1$, where $I_{HQ}$ is a high-quality ground truth image.
\vspace{-5pt}

\subsection{Fair Training}
\label{fair_training}
In this section, we identify two unfair problems in existing studies, and present our training strategies to resolve each problem.

\begin{figure}[t]
    \centering
    \includegraphics[width=\linewidth]{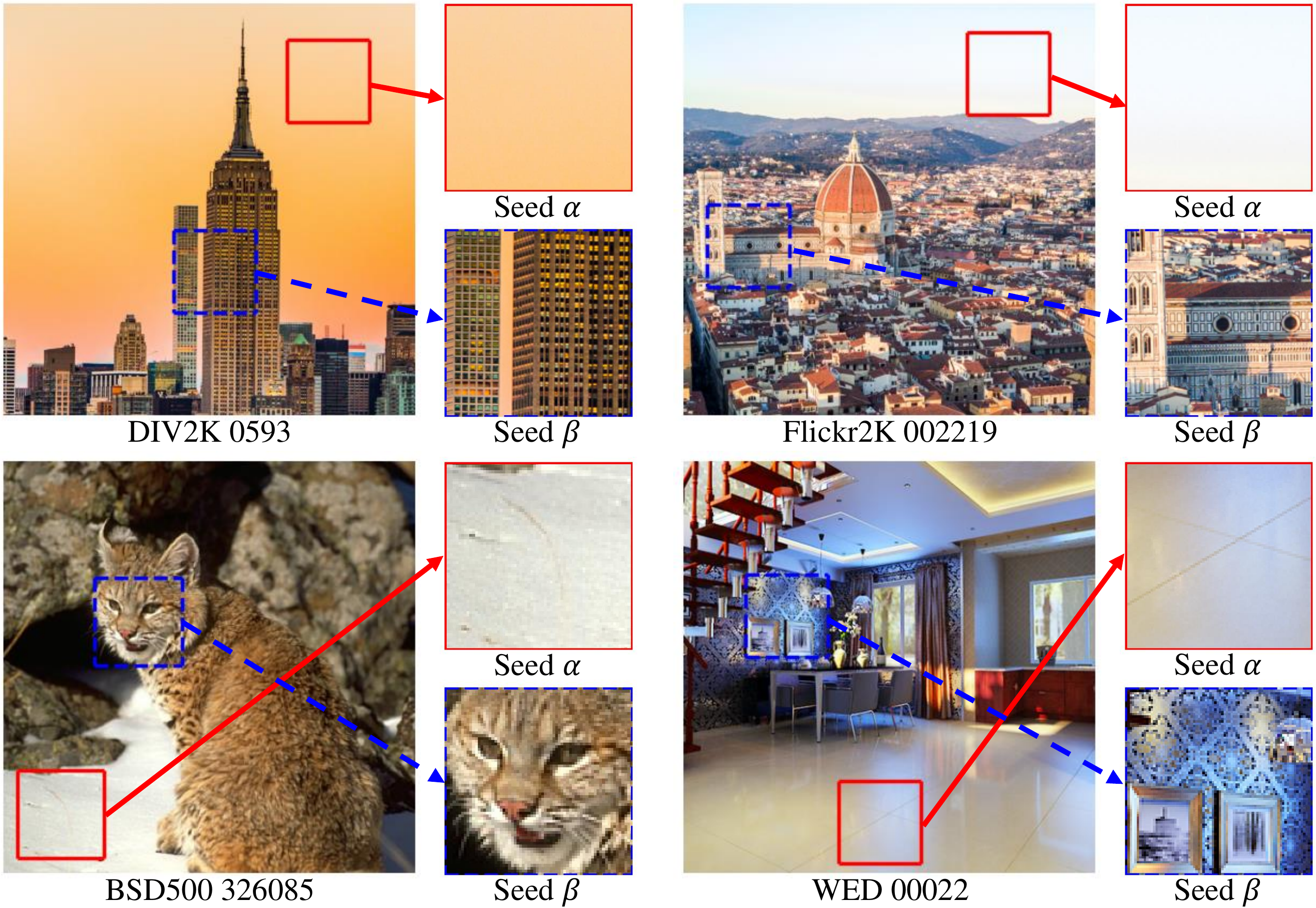}
    \caption{Examples of randomly cropped patches according to a random seed $\alpha$ or $\beta$. The random seed $\beta$ can select more the regions challenging to recover than $\alpha$. In the extreme cases, $\alpha$ leads to lower performances, as in Table~\ref{tab_seed}.}
    \label{fig_random_patch}
    \vspace{-10pt}
\end{figure}

\begin{figure}[t]
    \centering
    \includegraphics[width=0.8\linewidth]{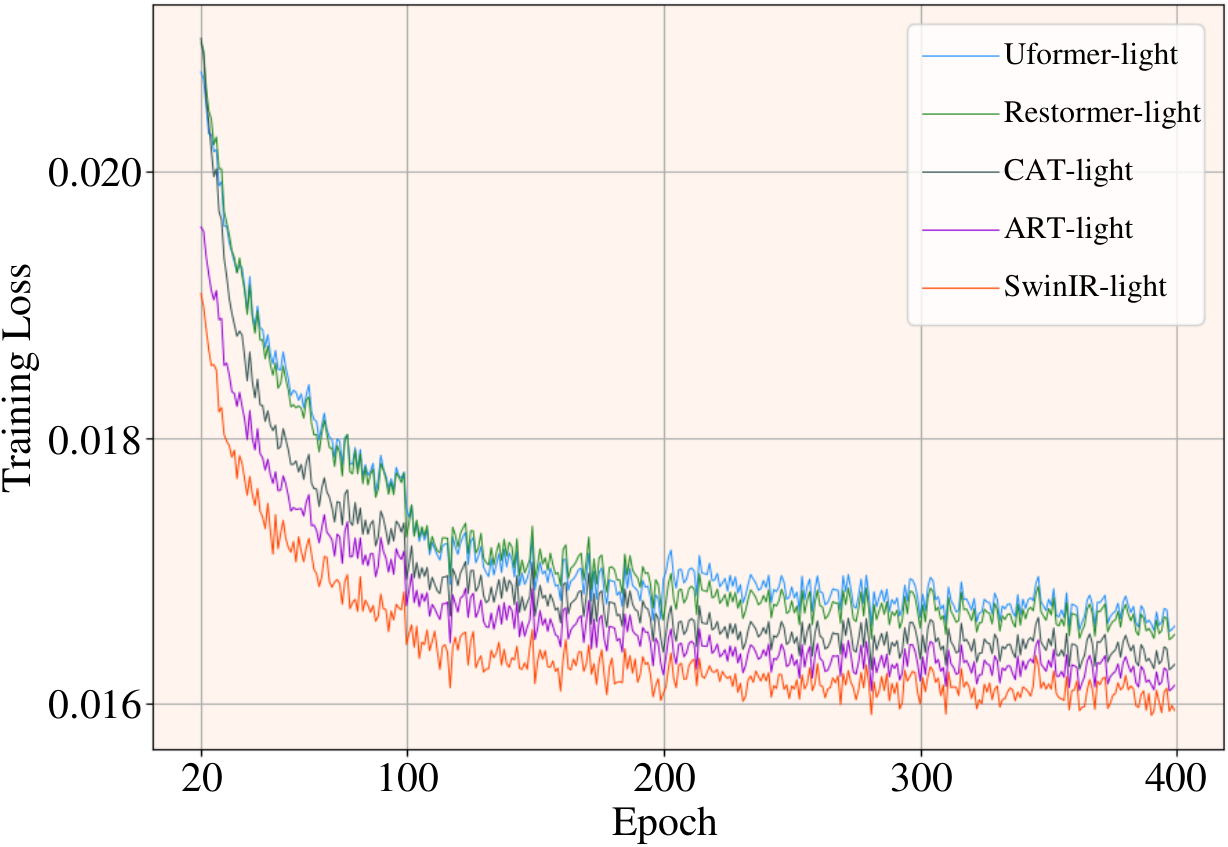}
    \caption{Trends of training loss of each model. The isomorphic movements across all models along each epoch means that the identical patches are used at a certain iteration. Note that the training loss of NGswin and ELAN-light are compared in Section~\ref{unstable} to describe the instability of ELAN-light.}
    \label{fig_trainloss}
\end{figure}

\textbf{Foremost}, most recent denoising studies have trained their models on randomly cropped patches from training images~\cite{liang2021swinir, wang2022uformer, zamir2022restormer, zhang2022accurate, zheng2022cross, xiaostochastic}, because the resolution of the original image is too high to process with current hardware.
However, as opposite to conventional wisdom that numerous trials always lead to almost identical results, we discover that the areas randomly cropped from training data hugely influence the denoising performances.
Even if existing studies have striven to compare models fairly, it was unfair at least for denoising task.
For example, assume that an image $I_{LQ}$ is used for training the networks at a iteration, as illustrated in Figure~\ref{fig_random_patch}.
While one random seed $\alpha$ crops a patch that is relatively easy to recover (\emph{e.g.}, background sky or ground), another random seed $\beta$ crops a patch that is challenging to restore (\emph{e.g.}, complex pattern or texture)~\cite{lin2022revisiting, gu2021interpreting}.
Even when the learned network architecture is the same, a network using random seed $\beta$ (or $\alpha$) shows better performances than $\alpha$ (or $\beta$) (Table~\ref{tab_seed}).
We, therefore, struggle to control every randomness that can appear during training.
The same random patch from a training image is guaranteed to be chosen through all networks at a certain iteration.
The identical data augmentation (see Section~\ref{setup}) is also applied at that iteration.
We cross-check whether the same patches are really used for training.
Figure~\ref{fig_trainloss} reveals that the fair training is realized.
The isomorphic movement of loss of every network means that identical data points are used for training the different models.

In implement, the mini-batch size and the number of GPUs affect the randomly selected patches or augmentation parameters.
Some models, such as SwinIR-light and ART-light, require more GPU memory than the others, which result in a smaller batch size or more GPUs.
It causes the random patches and augmentation to alter.
Therefore, we record the vertical and horizontal start points of cropped areas, as well as the random augmentation parameters (flip and roation), at each iteration while training a model.
This information is loaded when training the others.

\textbf{Next}, the common method to generate random noise is to exploit additive white Gaussian noise (AWGN).
This follows an assumption that Gaussian distribution can approximate the distribution of real-world unknown noise~\cite{li2021lightweight}.
Given a high-quality image $I_{HQ}$, a low-quality noisy image $I_{LQ}$ can be produced as follows:
\begin{equation}
    \label{eq_noise}
    \begin{aligned}
        I_{LQ} = I_{HQ} + \mathcal{S}, \mathcal{S}\sim\mathcal{N}(0, \sigma^2),
    \end{aligned}
\end{equation}
where $\mathcal{S}$ denotes a noise term and $\sigma^2$ indicates the variance of Gaussian distribution $\mathcal{N}$.
$\sigma$ determines noise level, \emph{i.e.}, the larger $\sigma$ adds more noise.
While some studies use a constant $\sigma$ for training each independent model~\cite{zhang2019residual, zhang2020residual, chen2021pre, liang2021swinir}, others utilize a blind $\sigma$ to construct a single model~\cite{zhang2017beyond, zhang2017learning, zhang2018ffdnet, zhang2021plug}.
The latter is worse at restoring a specific $\sigma$ the former chooses.
In contrast, the former is bad at recovering noisy images from the other $\sigma$ values.
Because of this difference, it is unfair to compare the former and latter directly.
Thus, we get the low-quality noisy images by adding Gaussian noise with blind $\sigma$ (sampled uniformly between 0 and 50), and train all Transformers following this rule.

\section{Experiments}

\subsection{Experimental Setup}
\label{setup}
We implemented all works using PyTorch~\cite{paszke2019pytorch} on 2 NVIDIA GeForce RTX 4090 GPUs, including the model configurations, training, and evaluation procedures.

\noindent
\textbf{Training.}
Following previous works~\cite{liang2021swinir, zamir2022restormer,zhang2022accurate}, we used a merged dataset DFBW including 8,594 high-quality images (800 DIV2K~\cite{agustsson2017ntire}, 2,650 Flickr2K~\cite{timofte2017ntire}, 400 BSD500~\cite{arbelaez2010contour}, and 4,744 WED~\cite{ma2016waterloo}).
The training process lasted for 400 epochs.
As previously mentioned, a blind Gaussian noise was added to a high-quality image.
Moreover, we employed progressive learning following Restormer~\cite{zamir2022restormer}.
The patch size for random cropping was initialized as 64$\times$64 (batch size: 64) and then increased to 96$\times$96 (batch size: 32) and 128$\times$128 (batch size: 16) after 100 and 200 epochs, respectively.
As emphasized in Section~\ref{fair_training}, a random patch at a certain iteration was all the same for all models.
After random cropping, we augmented the data by random horizontal flipping and rotation ($90^{\circ}$, $180^{\circ}$, $270^{\circ}$).
The learning rate was initialized as $0.0004$, which is halved after $\{200, 300, 350, 375\}$ epochs.
For the first 20 epochs, there was warmup phase~\cite{goyal2017accurate} that linearly increased the learning rate from $0.0$ to $0.0004$.
We used Adam~\cite{kingma2014adam} optimizer.

\noindent
\textbf{Evaluation.}
We reported PSNR (dB) and SSIM~\cite{wang2004image} on the standard benchmark test datasets as metrics.
The test sets for color DN include CBSD68~\cite{martin2001database}, Kodak24~\cite{franzen1999kodak}, McMaster~\cite{zhang2011color}, and Urban100~\cite{huang2015single}.
The performances on Set12~\cite{zhang2017beyond}, BSD68~\cite{martin2001database}, and Urban100~\cite{huang2015single} for grayscale DN were evaluated.
The noise levels $\sigma$ of evaluation were 15, 25, and 50.

\subsection{Main Results of Baselines}
\label{main_results}
As shown in Table~\ref{tab_color_base}, we compare our fairly trained lightweight Transformer baselines for color blind Gaussian denoising (DN).
We witnessed two interesting points in this table.

In terms of the \textbf{original task} of each model, the networks from lightweight super-resolution (SR) field generally perform better than the counterparts stemming from large DN.
This differences result from a reason that the methods from lightweight SR were already designed to perform efficiently.
It implies that lightening deep neural networks is beyond simply reducing the number of parameters.
Therefore, we discuss this issue in Section~\ref{analysis} to provide some considerations and insights when designing a effective lightweight network.
Although not covered in this work, more sophisticated skills, such as quantization~\cite{li2020pams, wang2021fully, liu2021super, hong2022daq, hong2022cadyq} or network pruning~\cite{jiang2021learning, zhang2021learning, zhan2021achieving}, may be also considered.

\begin{table*}[t]
    \centering
    \caption{The results of our baselines for blind Gaussian denoising. We ensure the entirely identical settings for training and testing. The first, second, and third best performances are in \textcolor{red}{red}, \textcolor{blue}{blue}, and \textcolor{mygreen}{green}. \enquote{OOM} represents Out-Of-Memory.}
    \label{tab_baseline}
    \begin{subtable}[h]{0.49\linewidth}
    \caption{Color blind Gaussian denoising baselines.}
    \label{tab_color_base}
    \centering
    \resizebox{\linewidth}{!}
    {
    \begin{tabular}{|c|c|c|c|c|c|c|c|c|c|c|}
    \hline
    \multirow{2}{*}{Model} & \multirow{2}{*}{\#Params} & \multirow{2}{*}{$\sigma$} & \multicolumn{2}{c|}{CBSD68~\cite{martin2001database}} & \multicolumn{2}{c|}{Kodak24~\cite{franzen1999kodak}} & \multicolumn{2}{c|}{McMaster~\cite{zhang2011color}} & \multicolumn{2}{c|}{Urban100~\cite{huang2015single}} \\ \cline{4-11}
    & & & PSNR & SSIM & PSNR & SSIM & PSNR & SSIM & PSNR & SSIM \\
    \hline
    SwinIR-light & 905K & \multirow{3}{*}{15} & \textcolor{red}{34.16} & \textcolor{blue}{0.9323} & \textcolor{red}{35.18} & \textcolor{red}{0.9269} & \textcolor{red}{35.23} & \textcolor{red}{0.9295} & \textcolor{red}{34.59} & \textcolor{red}{0.9478} \\
    ELAN-light & 616K & & 34.06 & 0.9312 & \textcolor{mygreen}{35.06} & \textcolor{mygreen}{0.9256} & 35.09 & 0.9277 & \textcolor{mygreen}{34.47} & \textcolor{mygreen}{0.9464} \\
    NGswin & 993K & & \textcolor{blue}{34.12} & \textcolor{red}{0.9324} & \textcolor{blue}{35.12} & \textcolor{blue}{0.9268} & \textcolor{blue}{35.17} & \textcolor{blue}{0.9294} & \textcolor{blue}{34.53} & \textcolor{blue}{0.9476} \\
    \hdashline
    Restormer-light & 1,054K & \multirow{4}{*}{15} & 33.99 & 0.9311 & 34.86 & 0.9244 & 34.69 & 0.9229 & 34.00 & 0.9439 \\
    Uformer-light & 1,084K & & 34.02 & 0.9310 & 34.91 & 0.9246 & 34.81 & 0.9241 & 34.04 & 0.9442 \\
    CAT-light & 1,042K & & 34.01 & 0.9304 & 34.90 & 0.9237 & 34.83 & 0.9247 & OOM & OOM \\
    ART-light & 1,084K & & \textcolor{mygreen}{34.08} & \textcolor{mygreen}{0.9315} & 35.00 & 0.9251 & \textcolor{mygreen}{35.10} & \textcolor{mygreen}{0.9282} & OOM & OOM \\
    \hline
    SwinIR-light & 905K & \multirow{3}{*}{25} & \textcolor{red}{31.50} & \textcolor{blue}{0.8883} & \textcolor{red}{32.69} & \textcolor{red}{0.8868} & \textcolor{red}{32.90} & \textcolor{blue}{0.8977} & \textcolor{red}{32.23} & \textcolor{red}{0.9222} \\
    ELAN-light & 616K & & 31.39 & 0.8864 & \textcolor{mygreen}{32.56} & \textcolor{mygreen}{0.8846} & \textcolor{mygreen}{32.76} & 0.8950 & \textcolor{mygreen}{32.09} & \textcolor{mygreen}{0.9198} \\
    NGswin & 993K & & \textcolor{blue}{31.44} & \textcolor{red}{0.8884} & \textcolor{blue}{32.61} & \textcolor{blue}{0.8865} & \textcolor{blue}{32.82} & \textcolor{red}{0.8978} & \textcolor{blue}{32.13} & \textcolor{blue}{0.9215} \\
    \hdashline
    Restormer-light & 1,054K & \multirow{4}{*}{25} & 31.33 & 0.8865 & 32.38 & 0.8833 & 32.44 & 0.8905 & 31.60 & 0.9161 \\
    Uformer-light & 1,084K & & 31.38 & \textcolor{mygreen}{0.8866} & 32.44 & 0.8836 & 32.59 & 0.8922 & 31.67 & 0.9165 \\
    CAT-light & 1,042K & & 31.37 & 0.8855 & 32.43 & 0.8822 & 32.58 & 0.8928 & OOM & OOM \\
    ART-light & 1,084K & & \textcolor{mygreen}{31.40} & 0.8864 & 32.49 & 0.8833 & 32.74 & \textcolor{mygreen}{0.8956} & OOM & OOM \\
    \hline
    SwinIR-light & 905K & \multirow{3}{*}{50} & \textcolor{red}{28.22} & \textcolor{blue}{0.8006} & \textcolor{red}{29.54} & \textcolor{red}{0.8089} & \textcolor{red}{29.71} & \textcolor{red}{0.8339} & \textcolor{red}{28.89} & \textcolor{red}{0.8658} \\
    ELAN-light & 616K & & 28.07 & 0.7957 & \textcolor{mygreen}{29.35} & 0.8028 & \textcolor{mygreen}{29.51} & 0.8277 & \textcolor{mygreen}{28.67} & \textcolor{mygreen}{0.8596} \\
    NGswin & 993K & & \textcolor{blue}{28.13} & \textcolor{red}{0.8011} & \textcolor{blue}{29.42} & \textcolor{blue}{0.8087} & \textcolor{blue}{29.59} & \textcolor{red}{0.8339} & \textcolor{blue}{28.75} & \textcolor{blue}{0.8644} \\
    \hdashline
    Restormer-light & 1,054K & \multirow{4}{*}{50} & 28.04 & \textcolor{mygreen}{0.7974} & 29.19 & \textcolor{mygreen}{0.8034} & 29.31 & 0.8256 & 28.30 & 0.8559 \\
    Uformer-light & 1,084K & & \textcolor{mygreen}{28.11} & 0.7968 & 29.26 & 0.8020 & 29.46 & 0.8259 & 28.33 & 0.8551 \\
    CAT-light & 1,042K & & \textcolor{mygreen}{28.11} & 0.7960 & 29.29 & 0.8024 & 29.48 & \textcolor{blue}{0.8296} & OOM & OOM \\
    ART-light & 1,084K & & 28.08 & 0.7950 & 29.27 & 0.8000 & 29.48 & \textcolor{mygreen}{0.8279} & OOM & OOM \\
    \hline
    \end{tabular}
    }
    \end{subtable}
    \begin{subtable}[h]{0.49\linewidth}
    \vspace{5pt}
    \caption{Grayscale blind Gaussian denoising baselines.}
    \label{tab_gray_base}
    \centering
    \resizebox{0.82\linewidth}{!}
    {
    \begin{tabular}{|c|c|c|c|c|c|c|c|c|}
    \hline
    \multirow{2}{*}{Model} & \multirow{2}{*}{\#Params} & \multirow{2}{*}{$\sigma$} & \multicolumn{2}{c|}{Set12~\cite{zhang2017beyond}} & \multicolumn{2}{c|}{BSD68~\cite{martin2001database}} & \multicolumn{2}{c|}{Urban100~\cite{huang2015single}} \\ \cline{4-9}
    & & & PSNR & SSIM & PSNR & SSIM & PSNR & SSIM \\
    \hline
    SwinIR-light & 903K & \multirow{3}{*}{15} & \textcolor{red}{33.04} & \textcolor{blue}{0.9052} & \textcolor{blue}{31.78} & \textcolor{red}{0.8926} & \textcolor{red}{33.04} & \textcolor{red}{0.9317} \\
    ELAN-light & 613K & & \textcolor{blue}{33.01} & \textcolor{mygreen}{0.9044} & 31.74 & 0.8910 & \textcolor{mygreen}{32.97} & 0.9299 \\
    NGswin & 991K & & \textcolor{red}{33.04} & \textcolor{red}{0.9055} & \textcolor{blue}{31.78} & \textcolor{blue}{0.8927} & \textcolor{blue}{32.99} & \textcolor{blue}{0.9314} \\
    \hdashline
    Restormer-light & 1,053K & \multirow{4}{*}{15} & 32.93 & 0.9039 & \textcolor{mygreen}{31.76} & \textcolor{mygreen}{0.8922} & 32.81 & \textcolor{mygreen}{0.9306} \\
    Uformer-light & 1,084K & & 32.88 & 0.9034 & 31.70 & 0.8910 & 32.66 & 0.9286 \\
    CAT-light & 1,041K & & 32.91 & 0.9021 & \textcolor{red}{31.89} & 0.8913 & OOM & OOM \\
    ART-light & 1,082K & & 32.93 & 0.9023 & 31.73 & 0.8911 & OOM & OOM \\
    \hline
    SwinIR-light & 903K & \multirow{3}{*}{25} & \textcolor{red}{30.67} & \textcolor{blue}{0.8669} & \textcolor{mygreen}{29.32} & \textcolor{blue}{0.8325} & \textcolor{red}{30.52} & \textcolor{red}{0.8963} \\
    ELAN-light & 613K & & \textcolor{blue}{30.65} & \textcolor{mygreen}{0.8665} & 29.29 & 0.8304 & \textcolor{blue}{30.46} & 0.8950 \\
    NGswin & 991K & & \textcolor{blue}{30.65} & \textcolor{red}{0.8671} & \textcolor{blue}{29.33} & \textcolor{mygreen}{0.8324} & \textcolor{blue}{30.46} & \textcolor{blue}{0.8961} \\
    \hdashline
    Restormer-light & 1,053K & \multirow{4}{*}{25} & 30.60 & 0.8659 & \textcolor{mygreen}{29.32} & 0.8322 & 30.32 & \textcolor{mygreen}{0.8952} \\
    Uformer-light & 1,084K & & 30.57 & 0.8650 & 29.26 & 0.8303 & 30.21 & 0.8929 \\
    CAT-light & 1,041K & & 30.60 & 0.8641 & \textcolor{red}{29.47} & \textcolor{red}{0.8330} & OOM & OOM \\
    ART-light & 1,082K & & 30.52 & 0.8620 & 29.25 & 0.8285 & OOM & OOM \\
    \hline
    SwinIR-light & 903K & \multirow{3}{*}{50} & \textcolor{red}{27.50} & \textcolor{red}{0.7966} & \textcolor{mygreen}{26.35} & \textcolor{blue}{0.7299} & \textcolor{red}{27.01} & \textcolor{blue}{0.8190} \\
    ELAN-light & 613K & & 27.46 & 0.7959 & 26.33 & 0.7269 & \textcolor{blue}{26.93} & 0.8172 \\
    NGswin & 991K & & 27.42 & \textcolor{blue}{0.7961} & \textcolor{blue}{26.38} & \textcolor{mygreen}{0.7298} & \textcolor{blue}{26.96} & \textcolor{red}{0.8192} \\
    \hdashline
    Restormer-light & 1,053K & \multirow{4}{*}{50} & \textcolor{mygreen}{27.48} & \textcolor{mygreen}{0.7960} & \textcolor{blue}{26.38} & 0.7285 & 26.92 & \textcolor{blue}{0.8190} \\
    Uformer-light & 1,084K & & 27.43 & 0.7934 & 26.33 & 0.7262 & 26.81 & 0.8154 \\
    CAT-light & 1,041K & & \textcolor{blue}{27.49} & 0.7935 & \textcolor{red}{26.52} & \textcolor{red}{0.7333} & OOM & OOM \\
    ART-light & 1,082K & & 27.26 & 0.7856 & 26.25 & 0.7194 & OOM & OOM \\
    \hline
    \end{tabular}
    }
    \end{subtable}
\end{table*}

\begin{table}[t]
    \centering
    \caption{Study on randomness. The random seed $\alpha$ is the same as what our baselines follow. Another seed $\beta$ differs from $\alpha$. The results marked as a same seed mean that the identical patches and corresponding augmentation are used at a certain iteration. PSNR / SSIM are evaluated with $\sigma=50$.} 
    \label{tab_seed}
    \resizebox{\linewidth}{!}
    {\begin{tabular}{|c|c|c|c|c|c|}
    \hline
    Method & Seed & CBSD68~\cite{martin2001database} & Kodak24~\cite{franzen1999kodak} & McMaster~\cite{zhang2011color} & Urban100~\cite{huang2015single} \\
    \hline
    \multirow{2}{*}{ELAN-light} & $\alpha$ & 28.07 / 0.7957 & 29.35 / 0.8028 & 29.51 / 0.8277 & 28.67 / 0.8596 \\
    & $\beta$ & \textbf{28.20} / \textbf{0.8002} & \textbf{29.49} / \textbf{0.8087} & \textbf{29.65} / \textbf{0.8338} & \textbf{28.85} / \textbf{0.8651} \\
    \hline
    \multirow{2}{*}{NGswin} & $\alpha$ & 28.13 / 0.8011 & 29.42 / 0.8087 & 29.59 / 0.8339 & 28.75 / 0.8644 \\
    & $\beta$ & \textbf{28.27} / \textbf{0.8027} & \textbf{29.58} / \textbf{0.8114} & \textbf{29.75} / \textbf{0.8362} & \textbf{28.90} / \textbf{0.8671} \\
    \hline
    \multirow{2}{*}{Restormer-light} & $\alpha$ & 28.04 / \textbf{0.7974} & 29.19 / \textbf{0.8034} & 29.31 / \textbf{0.8256} & \textbf{28.30} / \textbf{0.8559} \\
    & $\beta$ & \textbf{28.11} / 0.7951 & \textbf{29.25} / 0.8008 & \textbf{29.35} / 0.8248 & 28.28 / 0.8533 \\
    \hline
    \multirow{2}{*}{Uformer-light} & $\alpha$ & 28.11 / 0.7968 & 29.26 / 0.8020 & 29.46 / 0.8259 & 28.33 / 0.8551 \\
    & $\beta$ & \textbf{28.12} / \textbf{0.7986} & \textbf{29.34} / \textbf{0.8051} & \textbf{29.51} / \textbf{0.8299} & \textbf{28.44} / \textbf{0.8591} \\
    \hline
    \end{tabular}}
    \vspace{-5pt}
\end{table}

Next, with respect to the \textbf{network architecture}, non-hierarchical structure (please remind Table~\ref{tab_summary}) results in better performances on lower noise level.
Non-hierarchical ART-light performs the best among the networks from large DN (below dashline) on $\sigma=15,25$.
As demonstrated in~\cite{choi2022n}, this is because reconstruction of high quality image by utilizing higher resolution features is more straightforward than by handling smaller features.
But situation changes when recovering the highly distorted images ($\sigma=50$).
ART-light only shows similar results to Uformer-light and CAT-light.
Other algorithms of self-attention or FFN arranged in Table~\ref{tab_summary} affected this challenging task.
Meanwhile, NGswin seems to overcome the issue of hierarchical network by several crucial components designed efficiently (see Section~\ref{hier}).
In addition, Restormer-light shows the low reconstruction performances.
It employs channel self-attention to capture global dependency of every pixel instead of local spatial self-attention adopted in the other baselines.
While the large DN model (Restormer~\cite{zamir2022restormer}) achieved their goal by a number of parameters, Restormer-light lacks at the capacity to consider sufficient spatial information due to parameter constraint (around one milion).
It is discussed in Section~\ref{chattn}.

Secondarily, we also provide lightweight Transformer baselines for grayscale blind Gaussian denoising in Table~\ref{tab_gray_base}.
The results were similar to color denoising.
Interestingly, however, CAT-light recorded outstanding results especially on BSD68 dataset.
From the result, we drew the possibility that a task- or dataset-oriented architecture can be designed intentionally.

\begin{figure*}[t]
    \centering
    \begin{tabular}{c}
        \includegraphics[width=0.9\linewidth]{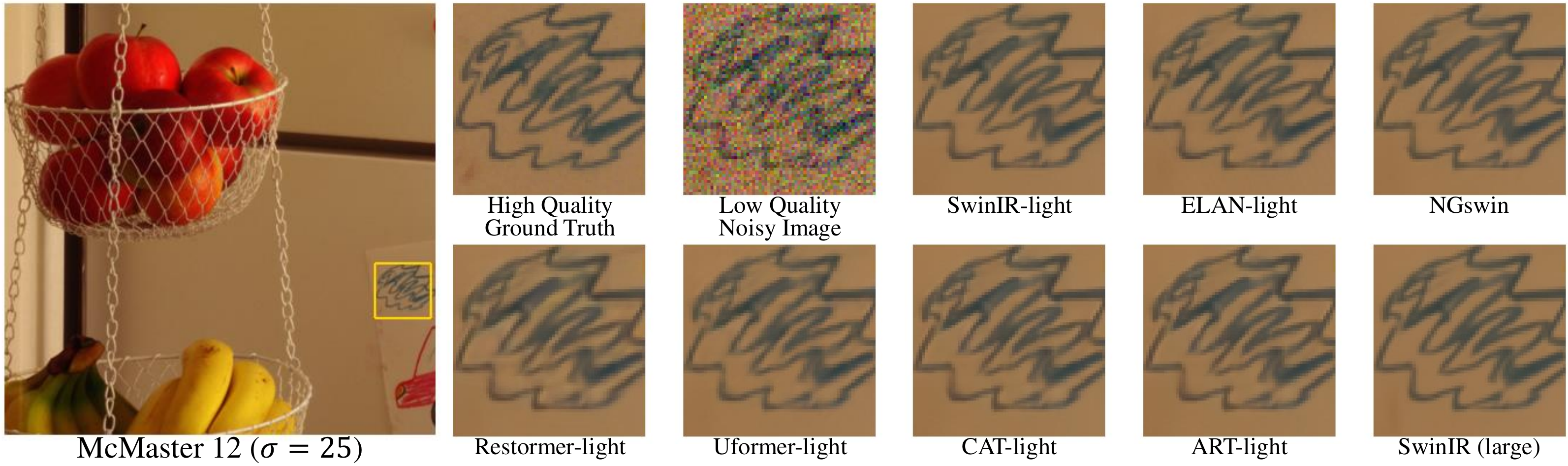} \\
        \includegraphics[width=0.9\linewidth]{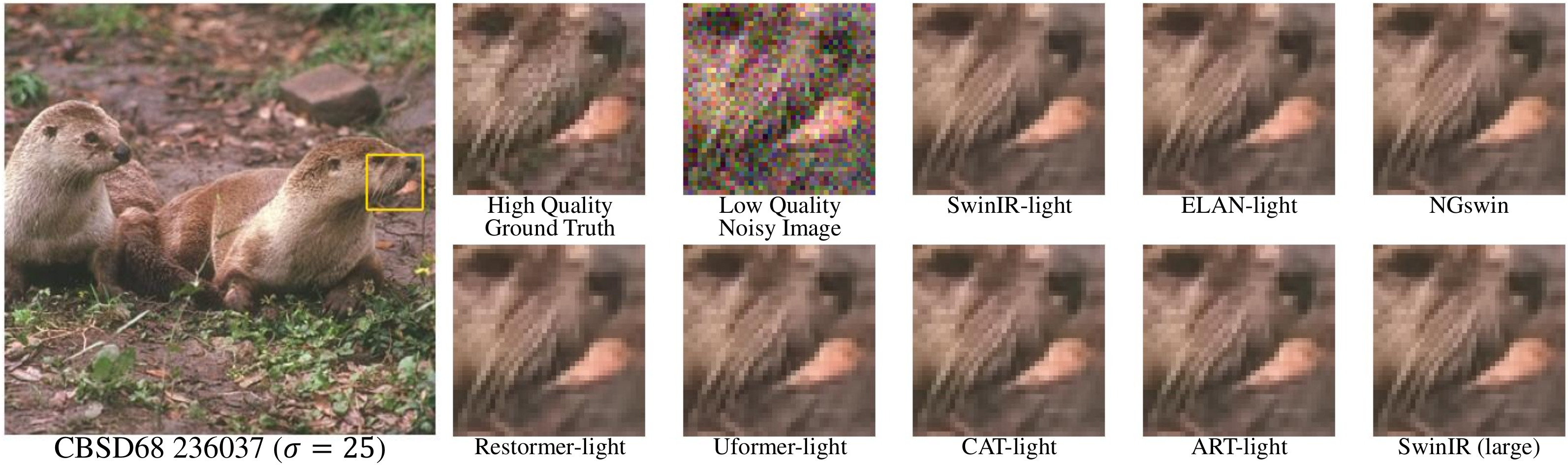} \\
    \end{tabular}
    \caption{The visual comparison of denoising results of our seven baseline Transformers and a large model. While the large SwinIR recovers degraded images the best, our baselines can generally produce the comparable results for human-perception with much fewer parameters.}
    \label{fig_viscomp}
\end{figure*}

The visual comparisons are supplied in Figure~\ref{fig_viscomp}.

\subsection{Analysis of Randomness}
\label{randomness}

\begin{table}[t]
    \centering
    \caption{Study on causal verification of random elements. \enquote{Seed for data} $x$ means the ares randomly cropped  follow the random seed $x$. If \enquote{seed for init.} is set to $x$, the weights are initialized using the random seed $x$. The underlined results are the same as Uformer-light using $\alpha$ or $\beta$ in Table~\ref{tab_seed}.}
    \label{tab_init}
    \resizebox{\linewidth}{!}
    {\begin{tabular}{|cc|cc|cc|cc|cc|}
    \hline
    Seed & Seed & \multicolumn{2}{c|}{CBSD68~\cite{martin2001database}} & \multicolumn{2}{c|}{Kodak24~\cite{franzen1999kodak}} & \multicolumn{2}{c|}{McMaster~\cite{zhang2011color}} & \multicolumn{2}{c|}{Urban100~\cite{huang2015single}} \\
    for data & for init. & PSNR & SSIM & PSNR & SSIM & PSNR & SSIM & PSNR & SSIM \\
    \hline
    \multirow{2}{*}{$\alpha$} & $\alpha$ & \underline{28.11} & \underline{0.7968} & \underline{29.26} & \underline{0.8020} & \underline{29.46} & \underline{0.8259} & \underline{28.33} & \underline{0.8551} \\
    & $\beta$ & 28.10 & 0.7963 & 29.30 & 0.8032 & 29.47 & 0.8263 & 28.32 & 0.8547 \\
    \hline
    \multirow{2}{*}{$\beta$} & $\alpha$ & 28.12 & 0.7987 & 29.35 & 0.8062 & 29.51 & 0.8299 & 28.43 & 0.8588 \\
    & $\beta$ & \underline{28.12} & \underline{0.7986} & \underline{29.34} & \underline{0.8051} & \underline{29.51} & \underline{0.8299} & \underline{28.44} & \underline{0.8591} \\
    \hline
    \end{tabular}}
    \vspace{-13pt}
\end{table}

As recorded in Table~\ref{tab_seed}, PSNR scores of all models on all datasets increased with a new seed, except for Restormer-light on Urban100.
SSIM values for all but Restormer-light also grew up.
For example, NGswin with new seed $\beta$ outperformed SwinIR-light using the original seed $\alpha$ (refer to Table~\ref{tab_baseline}).
In turn, ELAN-light with $\beta$ surpassed NGswin using $\alpha$.
It is demonstrated that a vast number of trials cannot solve problem of randomness at least in image denoising task.
Please note that those overall improved results are not attributed to a novel or smart approaches.
Rather, they proved accident selection of random seed gives more successful results.
By contrast to previous works that overlooked this problem, our attempt to fairly prepare the training patches and compare the models based on this fairness is compelling.
To support our findings, we verify the true cause of these results by comparing the results from randomly cropped data and randomly initialized weights in Table~\ref{tab_init}.
The latter could not make relatively meaningful differences when randomly cropped patches are maintained as the same at a certain iteration.
As a result, it is necessary to consider and control the training data resulting from randomness for truly fair comparison.

\subsection{Empirical Analysis of Components}
\label{analysis}
\subsubsection{Hierarchical Structure}
\label{hier}

The hierarchical structures have been widely employed in the general image restoration (IR) tasks for the network efficiency~\cite{zamir2022restormer,wang2022uformer,zheng2022cross,choi2022n,zhang2021plug,ji2022xydeblur}.
Among our LWDN Transformer baselines, NGswin, Restormer-light, Uformer-light, and CAT-light utilize this U-Net~\cite{ronneberger2015u} based architectures (recall Table~\ref{tab_summary}).
However, the layers taking and producing lower-resolution features lose the spatial details of high-frequency information~\cite{choi2022n}.
Considering the degradation in other IR tasks (\emph{e.g.}, deraining, demosaicing) follows a relatively homogeneous pattern, preserving high-frequency details is particularly crucial in denoising task to recover edges and textures destroyed by heterogeneous random noise.
Thus, the hierarchical denoiser tends to fall behind the non-hierarchical structures when the parameter budget is maintained similar.
The fact that the non-hierarchical SwinIR-light is the best baselines highlights the importance of this issue.
Although Restormer~\cite{zamir2022restormer}, Uformer~\cite{wang2022uformer}, and CAT~\cite{zheng2022cross} (\emph{i.e.}, large DN models) tried to overcome it by enlarging their model size, they suffered from too many parameters (26M, 51M, and 26M, respectively).
This strategy is not reasonable in lightweight IR tasks that extremely constrain the network size (around 1M parameters in this paper).
Nevertheless, a hierarchical NGswin stops the significant drop of the performances.
In that point we investigate the U-Net components that can compensate the drawbacks efficiently.
\begin{table}[t]
    \centering
    \caption{The differences of hierarchical LWDN Transformers.} 
    \label{tab_hier}
    \resizebox{\linewidth}{!}
    {\begin{tabular}{c|c|c|c}
    \hline
    Method & Encoder Connection & Bottleneck Input & Decoder Structure \\
    \hline
    Restormer-light & None & Last encoder output & Symmetric \\
    Uformer-light & None & Last encoder output & Symmetric \\
    CAT-light & None & Last encoder output & Symmetric \\
    \hline
    NGswin & Dense connection~\cite{huang2017densely} & Merged multi-scale encoder features & Asymmetric~\cite{he2022masked} \\
    \hline
    \end{tabular}}
\end{table}

In Table~\ref{tab_hier}, we contrast NGswin with the other hierarchical baselines in terms of the main layers of U-Net.
First, NGswin placed a dense connectivity~\cite{huang2017densely} between encoder layers, while there were not any specific connections in the others.
This cascading mechanism conveys the information of the previous layers efficiently~\cite{ahn2018fast}.
Second, an input to a bottleneck layer is also different.
After the encoder stages, NGswin introduces the bottleneck taking merged multi-scale features.
It is named as SCDP; pixel-Shuffle, Concatenation, Depthwise convolution, and Point-wise projection.
SCDP can enhance the performances with the negligible extra parameters.
Third, NGswin exploits an asymmetric single decoder that is smaller than the encoder.
It not only highly increases the network efficiency but also takes advantages of high-resolution features.

\begin{table}[t]
    \centering
    \caption{Ablation study on the hierarchical structure. The baseline is Uformer-light. \bm{$\varDelta$} calculates the gaps over the baseline. The additional components are accumulated.}
    \label{tab_hier_ablation}
    \resizebox{\linewidth}{!}
    {\begin{tabular}{l|c|c|cc|cc|cc|cc}
    \hline
    \multirow{2}{*}{Configuration} & \multirow{2}{*}{\#Params} & \multirow{2}{*}{$\sigma$} & \multicolumn{2}{c|}{CBSD68~\cite{martin2001database}} & \multicolumn{2}{c|}{Kodak24~\cite{franzen1999kodak}} & \multicolumn{2}{c|}{McMaster~\cite{zhang2011color}} & \multicolumn{2}{c}{Urban100~\cite{huang2015single}} \\
    \cline{4-11}
    & & & PSNR & $\bm{\varDelta}$ & PSNR & $\bm{\varDelta}$ & PSNR & $\bm{\varDelta}$ & PSNR & $\bm{\varDelta}$ \\
    \hline
    Baseline & 1,084K & \multirow{4}{*}{15} & 34.02 & - & 34.91 & - & 34.81 & - & 34.04 & - \\
    + Dense Connection & 1,093K & & 34.02 & \:\:0.00 & 34.92 & \textcolor{red}{+0.01} & 34.82 & \textcolor{red}{+0.01} & 34.03 & {-0.01} \\
    + Multi-scale Bottleneck & 1,020K & & 34.10 & \textcolor{red}{+0.08} & 35.04 & \textcolor{red}{+0.13} & 34.97 & \textcolor{red}{+0.16} & 34.22 & \textcolor{red}{+0.18} \\
    + Asymmetric Decoder & 544K & & 34.10 & \textcolor{red}{+0.08} & 35.07 & \textcolor{red}{+0.16} & 35.09 & \textcolor{red}{+0.28} & 34.40 & \textcolor{red}{+0.36} \\
    \hline
    Baseline & 1,084K & \multirow{4}{*}{25} & 31.38 & - & 32.44 & - & 32.59 & - & 31.67 & - \\
    + Dense Connection & 1,093K & & 31.37 & -0.01 & 32.46 & \textcolor{red}{+0.02} & 32.59 & \:\:0.00 & 31.67 & -0.01 \\
    + Multi-scale Bottleneck & 1,020K & & 31.47 & \textcolor{red}{+0.09} & 32.60 & \textcolor{red}{+0.16} & 32.74 & \textcolor{red}{+0.15} & 31.88 & \textcolor{red}{+0.21} \\
    + Asymmetric Decoder & 544K & & 31.44 & \textcolor{red}{+0.06} & 32.59 & \textcolor{red}{+0.15} & 32.78 & \textcolor{red}{+0.19} & 32.01 & \textcolor{red}{+0.34} \\
    \hline
    Baseline & 1,084K & \multirow{4}{*}{50} & 28.11 & - & 29.26 & - & 29.46 & - & 28.33 & - \\
    + Dense Connection & 1,093K & & 28.10 & -0.01 & 29.31 & \textcolor{red}{+0.05} & 29.46 & \:\:0.00 & 28.35 & \textcolor{red}{+0.02} \\
    + Multi-scale Bottleneck & 1,020K & & 28.21 & \textcolor{red}{+0.10} & 29.48 & \textcolor{red}{+0.22} & 29.62 & \textcolor{red}{+0.16} & 28.61 & \textcolor{red}{+0.28} \\
    + Asymmetric Decoder & 544K & & 28.16 & \textcolor{red}{+0.05} & 29.43 & \textcolor{red}{+0.17} & 29.59 & \textcolor{red}{+0.13} & 28.65 & \textcolor{red}{+0.32} \\
    \hline
    \end{tabular}}
\end{table}

As shown in Table~\ref{tab_hier_ablation}, we conduct an ablation study applying those robust U-shaped components to Uformer-light, to inspect the potential of the hierarchical structures.
First of all, the features from the shallow module and each encoder layer are densely connected.
The performances slightly gain with a few additional parameters.
Next, we replaced the plain bottleneck with a modified SCDP.
We transformed some steps in SCDP of the original paper~\cite{choi2022n} due to the fundamental structural differences between NGswin and Uformer-light.
As this bottleneck only took the features before downsizing (\emph{i.e.}, the direct outputs from each encoder level), the 3rd downsizing layer was no more required.
Therefore, we could reduce the parameters but further enhance reconstruction accuracy.
The performances of enhanced Uformer-light were comparable to NGswin and ELAN-light (refer to Table~\ref{tab_baseline}).
Finally, we changed a symmetric decoder into an asymmetric one.
The three decoder levels were fused into one levels, which allows more encoder layers to be included.
The network depth shifts from $[2, 4, 2, 2, 2, 4, 2]$ to $[4, 4, 2, 2, 8]$.
Despite the deeper depth, removing existing decoders that took quite large channels enabled the number of parameters to be almost halved compared to the baseline.
This transformation also improved the performances.
It is demonstrated that the lightweight hierarchical network has the potential to progress.


\subsubsection{Spatial \textit{vs.} Channel Self-Attention}
\label{chattn}

It is ideal to involve every pixel of the feature maps in the spatial self-attention (SP-SA) computation as done in  ViT~\cite{dosovitskiy2020image} and IPT~\cite{chen2021pre}, but very high resolution of inputs for image restoration task leads to quadratic increase of time-complexity.
Thus, the origin~\cite{liang2021swinir,zhang2022efficient,choi2022n,wang2022uformer,zheng2022cross,zhang2022accurate} of our baselines employed the local window-based SP-SA except for Restormer~\cite{zamir2022restormer}.
Restormer utilized a channel self-attention (CH-SA) taking advantage of the global\footnote{In this section, the term \enquote{global} expresses that it involves all pixels of feature maps in computation of self-attention, not some pixels within a \enquote{local} window.} information, as local SP-SA is insufficient for considering global context.
The time-complexity\footnote{We omit other components proposed in each model, and softmax.} of typical local SP-SA and CH-SA are:
\begin{equation}
    \label{eq_complexity}
    \begin{aligned}
        & \Omega(\text{local SP-SA}) = 4H_iW_iC_i^2 + 2M^2H_iW_iC_i, \\
        & \Omega(\text{CH-SA}) = 4H_iW_iC_i^2 + 2H_iW_iC_i^2/L_i,
    \end{aligned}
\end{equation}
where $H_i$, $W_i$, and $C_i$, denote the height, width, and channels of feature maps in an $i$-\textit{th} Transformer block, and $M$ is a size of local window.
$L_i$ is the number of multi-heads.
CH-SA looks more efficient than                      SP-SA, as the main differences can be abbreviated as $M^2$ and $C_i/L_i$ in the second terms.

But there is a general trend that as the time complexity increases, so does the network capacity.
In other words, the capacity of CH-SA inversely proportional to $L_i$ means that more parallel multi-heads for attending to various spatial details from different perspectives~\cite{vaswani2017attention} reduces the network capacity.
In the models without parameter constraint (\emph{i.e.}, in larger models), this can be overcome by increasing the channels.
On the other hand, under a lightweight circumstance, the channels are highly reduced, which limits the increase of the parallel multi-heads in order to conserve capacity.
The inevitably limited (fewer) multi-heads, in turn, decrease the ability of attending to different parts of the input.
Correspondingly, CH-SA lacks the capability to capture and preserve semantic information in spatial dimension compared to SP-SA (Table~\ref{tab_color_base}).

\begin{figure}[t]
    \centering
    \includegraphics[width=\linewidth]{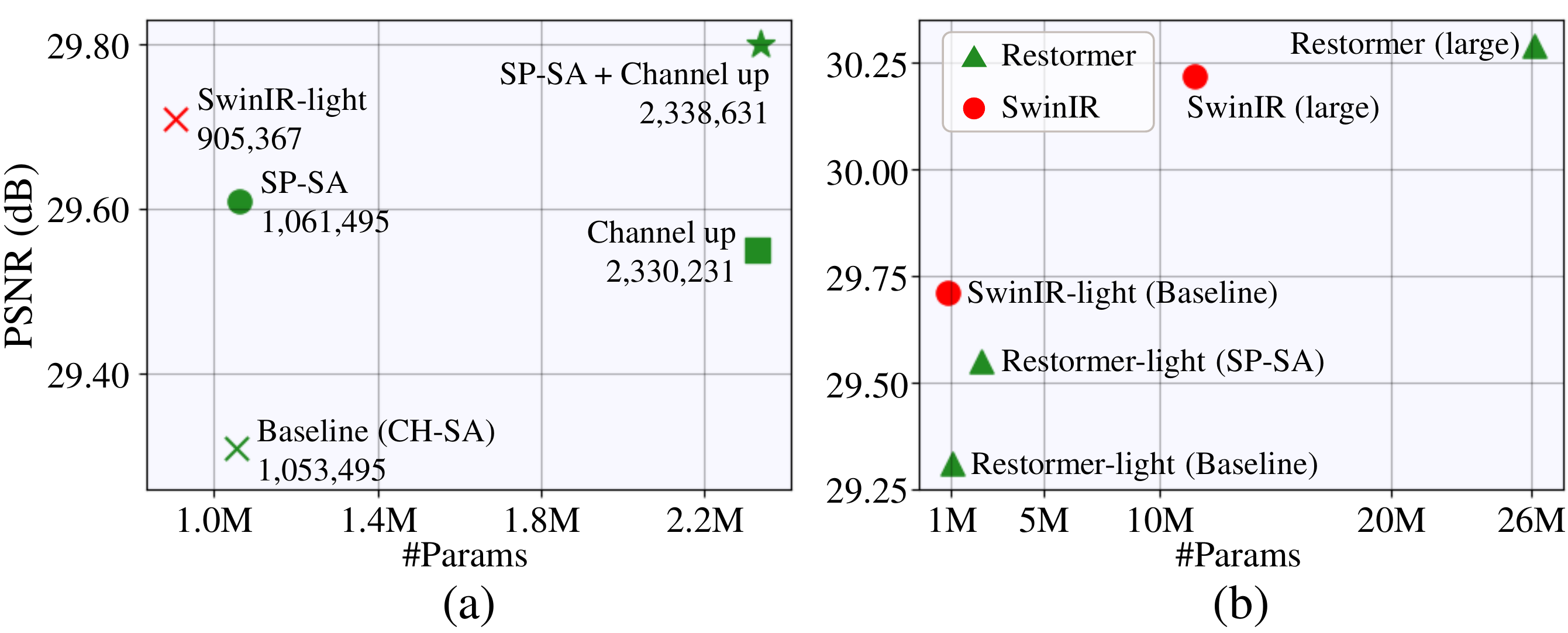}
    \caption{Ablation study on local spatial and channel self-attention. \textbf{(a)} The results of variants of Restormer-light. \textbf{(b)} The comparison with the large models. PSNR is evaluated on McMaster~\cite{zhang2011color} with $\sigma=50$.}
    \label{fig_chsa}
    \vspace{-13pt}
\end{figure}

To reinforce our claims, we conducted an ablation study in Figure~\ref{fig_chsa}\textcolor{red}{a}.
While the other structures or hyper-parameters were retained as the same of the baseline, we modified two components; the space of self-attention and the number of channels.
First, we tried to exploit global SP-SA following the original aim of Restoremer, but hardware was unable to endure massive complexity.
CH-SA of Restormer-light, therefore, was replaced with local square window-based SP-SA adopted in SwinIR-light, Uformer-light, and NGswin.
The result shows local SP-SA is superior over CH-SA under the lightweight condition.
The PSNR on McMaster dataset gains 0.3 dB with negligible extra parameters and time-complexity.
Second, we increased the channels while keeping CH-SA.
Despite a notable improvement with over twice the parameters, increasing channels did not meet SP-SA, which exposed the superiority of local SP-SA again.
Plus, when both modifications were applied, it
barely outperformed SwinIR-light with 2.58 times more parameters.
Finally, we compare the models in both large and lightweight size.
Figure~\ref{fig_chsa}\textcolor{red}{b} shows that CH-SA is effective without parameter constraints as mentioned before, whereas the effectiveness dwindles due to insufficient spatial comprehension in the lightweight field.

\subsubsection{Excessive weight sharing}
\label{unstable}

ELAN-light~\cite{zhang2022efficient} employed many weight sharing methods.
First, it proposed the accelerated self-attention, which shares the \textit{query} and \textit{key} in computation of self-attention (\emph{i.e.}, $Q = K$).
Second, once a shallower layer calculates the attention scores ($softmax(\frac{QK^T}{\sqrt{D}}), Q=K, D: dimension$), a consecutive layer shares them instead of separately producing them.
Third, ELAN-light employed shift-convolution~\cite{wu2018shift}, where several elements ,of which the original spatial locations and channels differ from each other, share the weight of a linear projection.

\begin{figure}[t]
    \centering
    \includegraphics[width=0.8\linewidth]{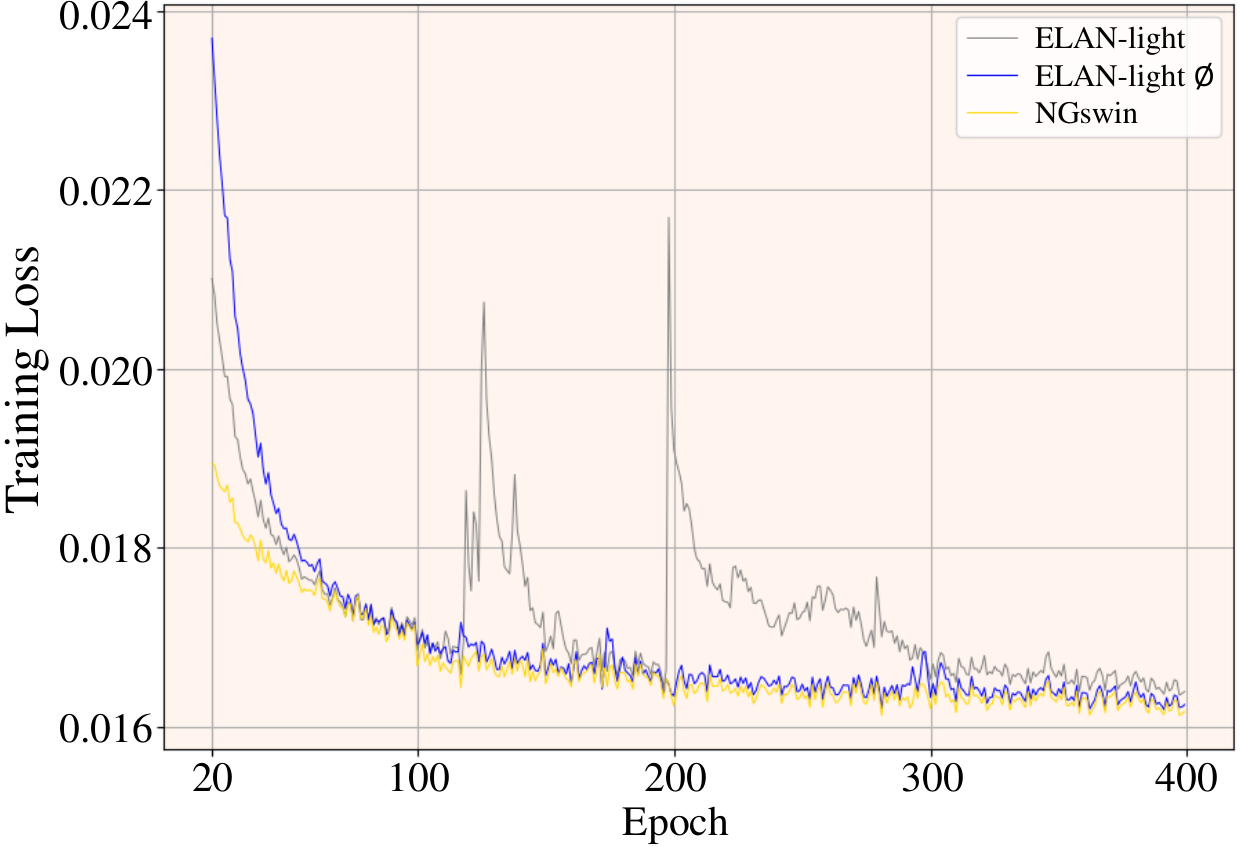}
    \caption{Trends of training loss. $\varnothing$ mark denotes removal of weight sharing in the model. The training of ELAN-light becomes unstable at some epochs. However, ELAN-light without weight sharing is trained stably.}
    \label{fig_elan_unstable}
\end{figure}

However, we figure out that the excessive weight sharing of this network leads to an unstable learning~\cite{xie2021weight}, as depicted in Figure~\ref{fig_elan_unstable}.
The training becomes stabilized when ELAN-light discards weight sharing methods.
The excessive weight sharing results in limited network flexibility and weak representation toward diverse inputs.
We hypothesize that those flaws may let a particular data point (an image patch) make the hypertrophied (overgrown) gradients during back-propagation.
This phenomenon causes the network parameters to diverge from optimal points in a moment, bringing out an abnormal loss.
Certainly, the mild weight sharing in a neural network is beneficial for some purposes, such as memory- and computation-efficiency.
Therefore, since the weight sharing leads to a trade-off between efficiency and flexibility, it is expected that future works aim to systematically find the optimal point of this trade-off.
Some regularization strategies, such as gradient clipping~\cite{pascanu2012understanding,pascanu2013difficulty}, or neural architecture search (NAS) methods~\cite{xie2021weight} can be helpful for handling this issue.

\begin{figure}[t]
    \centering
    \includegraphics[width=\linewidth]{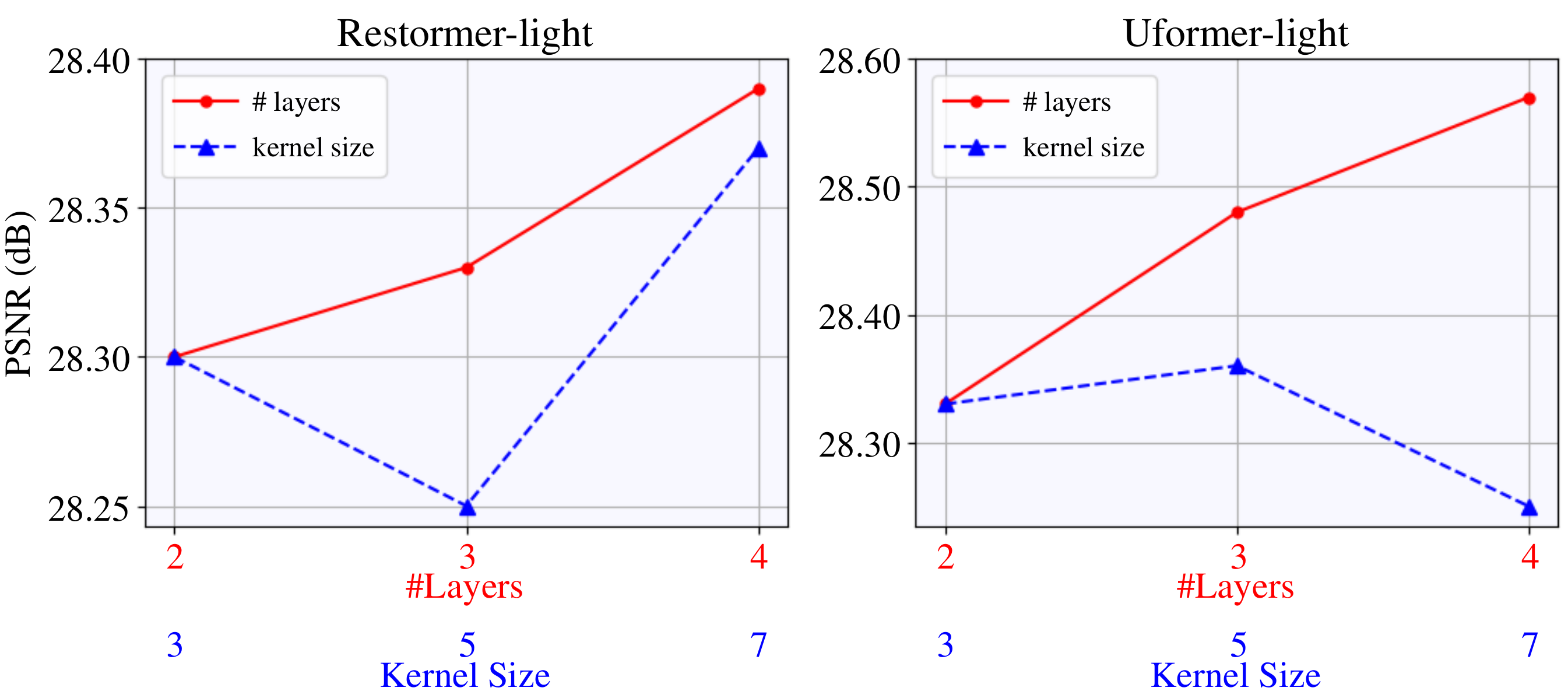}
    \caption{Study on tail variants. We increase the number of CNN layers or kernel size. PSNR is evaluated on Urban100~\cite{huang2015single} with $\sigma=50$.}
    \label{fig_tail_variants}
\end{figure}

\subsubsection{Still useful CNN}
\label{cnnstill}
Despite long-range dependency of the self-attention mechanism, the role of the meticulous composition of CNN is still relevant for image restoration tasks.
Unlike high-level vision tasks (\emph{e.g.}, classification, object detection), low-level tasks mainly aim to reconstruct each distorted pixel.
As this recovery process requires the information in the surrounding areas of each pixel~\cite{he2022masked, choi2022n,zheng2022cross}, CNN, which is conventionally good at extracting local features, is essential.
Figure~\ref{fig_tail_variants} visualizes the effect of variants of a reconstruction (tail) module, which is composed of only the convolutional layers.
In this experimental settings, we increased the number of convolutional layers or their kernel size.
The extra CNN layers added to the tail module outputted the same channels as the input features (a kernel size was fixed at $3\times3$).
When the kernel size increased, the number of layers was kept at 2.
As a result, the performance was proportional to the number of CNN in the tail module, while the kernel size followed case by case.

\subsubsection{A Supplement}
\begin{figure}[t]
    \centering
    \includegraphics[width=\linewidth]{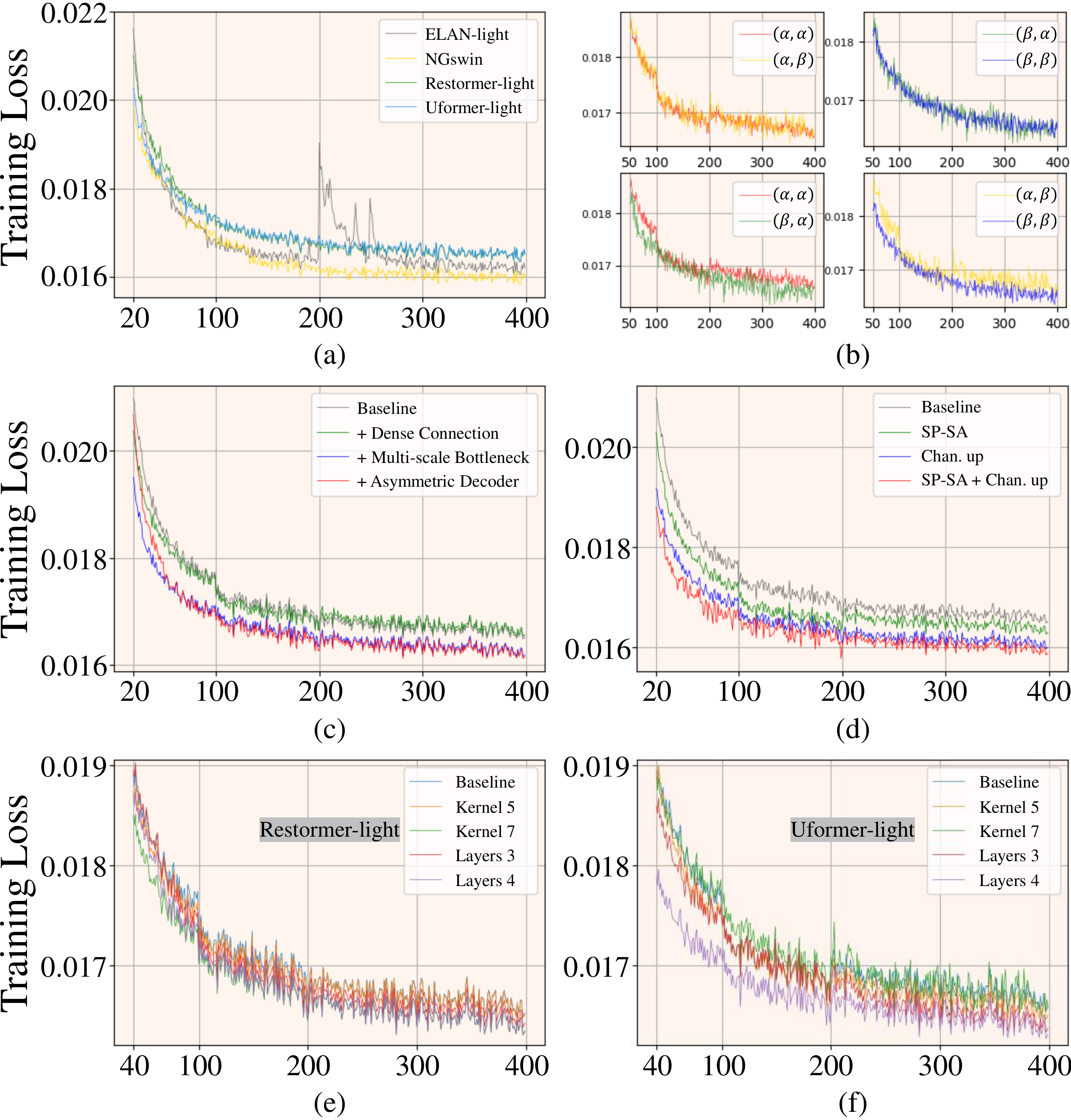}
    \caption{Training loss of all experiments in Section~\ref{analysis}. (a) Table~\ref{tab_seed}. (b) Table~\ref{tab_init}. (c) Table~\ref{tab_hier_ablation}. (d) Figure~\ref{fig_chsa}. (e), (f)  Figure~\ref{fig_tail_variants}.
    Note: the legends of (b) mean \texttt{(data seed, init seed)}, which reveals only the data seed can lead to similar trends of loss.}
    \label{fig_all_loss}
\end{figure}

In Figure~\ref{fig_all_loss}, we supply the training losses of all experiments in Section~\ref{analysis}.
Considering the similar movements of all of them, our crucial goal is achieved, the truly fair training.

\section{Conclusion}
This work presented seven Transformer baselines for lightweight denoising (LWDN), which has been unexplored until recently.
We aimed to control the randomness and train all models in a truly fair manner, because the patches randomly selected from a training image were found outstandingly influential in the recovery performances.
Based on our baselines, the empirical studies on different components delivered the considerations for LWDN with Transformers.
We verified the potential of hierarchical network to be further improved with the advanced elements, such as a dense connection, a multi-scale bottleneck, and an asymmetric decoder.
And it was proven more effective to utilize local window-based spatial self-attention in lightweight tasks rather than channel self-attention, unlike the models without parameter constraint.
Besides, excessive weight sharing caused the learning unstable, and the design of convolution was still relevant to denoising tasks.
In closing, we hope this work can encourage succeeding researchers to develop this field by using our baselines and findings.

\noindent
\textbf{Acknowledgements.}
This paper was supported by Institute of Information \& Communications Technology Planning \& Evaluation (IITP) grant (No.2022-0-00956) and Korea Health Industry Development
Institute (KHIDI) grant (No. H122C1983) funded by the Korea government (MSIT).

\bibliographystyle{ACM-Reference-Format}
\bibliography{sample-base}

\end{document}